\documentclass[letterpaper, 10 pt, journal, twoside]{ieeetran}

\IEEEoverridecommandlockouts            

\usepackage{cite}
\usepackage{amsmath,amssymb,amsfonts}
\usepackage{algorithmic}
\usepackage{algorithm}
\usepackage{graphicx}
\usepackage{textcomp}
\usepackage{xcolor}
\usepackage{url}
\usepackage{mathtools}
\usepackage{cite}

\DeclareMathOperator*{\argmax}{argmax} 
\DeclareMathOperator*{\argmin}{argmin} 

\def\BibTeX{{\rm B\kern-.05em{\sc i\kern-.025em b}\kern-.08em
    T\kern-.1667em\lower.7ex\hbox{E}\kern-.125emX}}

\author{Md. Ishat-E-Rabban, and Pratap Tokekar%
\thanks{Manuscript received: October, 15, 2020; Revised January, 10, 2021; Accepted February, 19, 2021.}
\thanks{This paper was recommended for publication by Editor M. Ani Hsieh upon evaluation of the Associate Editor and Reviewers' comments.
This work is supported by the National Science Foundation under Grant No. 1943368.} 
\thanks{Rabban and Tokekar are with Department of Computer Science, University of  Maryland  College  Park,  USA
        {\tt\footnotesize \{ier,tokekar\}@umd.edu}}%
\thanks{Digital Object Identifier (DOI): see top of this page.}
}

\title{Failure-Resilient Coverage Maximization with Multiple Robots}

\begin{document}

\maketitle

\begin{abstract}

The task of maximizing coverage using multiple robots has several applications such as surveillance, exploration, and environmental monitoring. A major challenge of deploying such multi-robot systems in a practical scenario is to ensure resilience against robot failures. A recent work~\cite{lz} introduced the 
\textit{Resilient Coverage Maximization} (\textit{RCM}) problem where the goal is to maximize a submodular coverage utility when the robots are subject to adversarial attacks or failures. The \textit{RCM} problem is known to be NP-hard.  In this paper, we propose two approximation algorithms for the \textit{RCM} problem, namely, the \textit{Ordered Greedy} (\textit{OrG}) and the \textit{Local Search} (\textit{LS}) algorithm. Both algorithms empirically outperform the state-of-the-art solution in terms of accuracy and running time. To demonstrate the effectiveness of our proposed solution, we empirically compare our proposed algorithms with the existing solution and a brute force optimal algorithm. We also perform a case study on the persistent monitoring problem to show the applicability of our proposed algorithms in a practical setting.


\end{abstract}


\section{Introduction}
\label{intro}

\IEEEPARstart{T}{asks} such as surveillance~\cite{rf2}, tracking~\cite{rf1}, and motion planning~\cite{rf7} can be formulated as an optimization problem that aims to maximize the coverage of a set of targets. These coverage maximization tasks can benefit from the use of multiple robots as opposed to a single robot. Although the advancements in robotic mobility, sensing, and communication technology have led to the use of multiple collaborating robots to support such tasks~\cite{rf4,rf5,rf6}, a major challenge for practical deployment of such multi-robot systems is to make the robots resilient to failures. For example, the robots may undergo adversarial attacks~\cite{rf13}, or the field-of-view of some robots may get occluded due to environmental hazards~\cite{rf14}, or the sensors may stop working due to technical malfunction~\cite{rf15}. In this paper, our goal is to devise coverage maximization algorithms that are resilient to such failures.

\begin{figure}[!ht]
\centering
\includegraphics[width=0.70\linewidth]{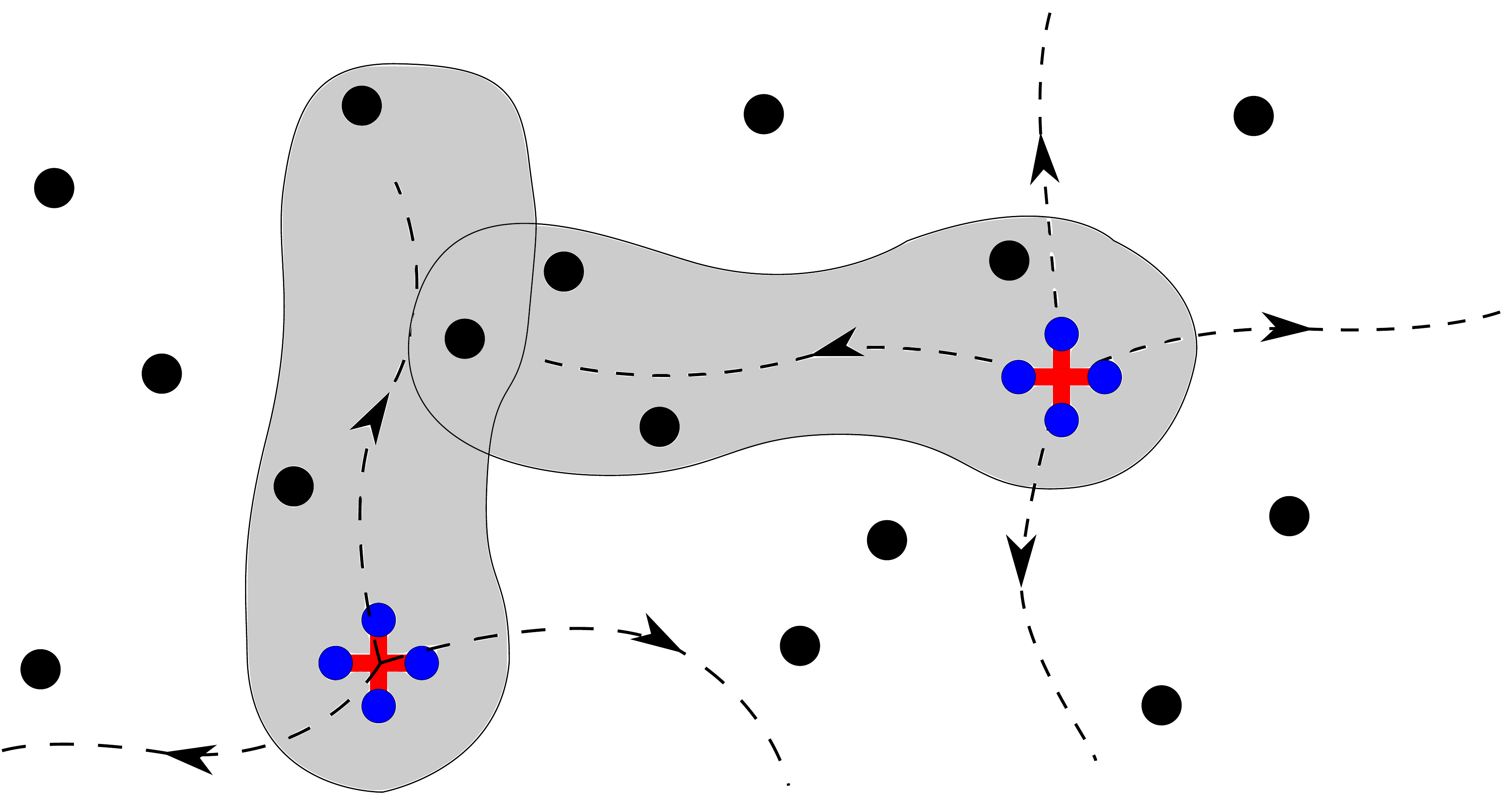}
\caption{Two robots are covering targets (black dots). The left robot has 3 available trajectories (dotted arrows) and the right one has 4. Coverage region of one trajectory of each robot is shown in gray. The highlighted trajectories of the left and right robots cover 3 and 4 targets respectively.}
\label{pdfig}
\end{figure}

Adversarial variants of combinatorial optimization problems have gained attention among the research community lately. For example, resilient resource allocation algorithms employ game theoretic strategy~\cite{rap1,rap2}, while adversarial coverage maximization algorithms use submodularity and greedy technique~\cite{rf10,rf16,rf18}. In a recent work, Zhou et al.~\cite{lz} introduced a new variant of the coverage maximization problem that takes into account the resilience of the multi-robot system. In this problem setup, a team of robots aim to cover a set of targets (Figure~\ref{pdfig}). For each robot, there is a set of candidate trajectories, one of which the robot will follow. The list of targets covered by each robot trajectory is provided. It is assumed that at most $\alpha$ robots may fail, but it is unknown which robots are going to fail. The objective of the problem is to select one trajectory for each robot such that the target coverage is maximized in the case of a worst-case failure of $\alpha$ robots. We call this problem \textit{Resilient Coverage Maximization} (\textit{RCM}) problem. The \textit{RCM} problem is known to be NP-hard~\cite{rf11}.
 

Building on a recent work of Tzoumas et al.~\cite{rf16} that studies generalized resilient optimization subject to matroid constraints, Zhou et al.~\cite{lz} presented an approximation algorithm for the \textit{RCM} problem that involves two phases. In the first phase, the algorithm determines the worst-case subset of $\alpha$ robots that could fail, and selects their trajectories. In the next phase, assuming that the robots selected in the first phase will actually fail, the rest of the robot trajectories are selected greedily such that, for each greedy selection, the marginal gain in target coverage is maximized. We call this algorithm the \textit{2 Phase Greedy} (\textit{2PG}) algorithm. The running time of the \textit{2PG} algorithm is O($P^2$), where $P$ is the sum of the number of candidate trajectories of all the robots.

In this paper, we propose two algorithms for the \textit{RCM} problem that outperform the \textit{2PG} algorithm both in terms of accuracy and running time. Here, by accuracy of a solution, we mean how much target coverage the solution achieves with respect to an optimal solution. Our proposed algorithms are called \textit{Ordered Greedy} (\textit{OrG}) algorithm and \textit{Local Search} (\textit{LS}) algorithm. 

The \textit{OrG} algorithm produces an ordering of the robots according to some sorting criteria, and greedily selects the trajectories of each robot sequentially according to the sorted order such that, for each robot, marginal increase in target coverage is maximized. The running time of the \textit{OrG} algorithm is O($P$). Experimental results show that the accuracy of the \textit{OrG} algorithm is slightly better than the \textit{2PG} algorithm, and it runs significantly faster than the \textit{2PG} algorithm.

In the \textit{LS} algorithm, we start with an initial solution of the \textit{RCM} problem. Then, in each iteration, we consider a set of \textit{neighbors} (to be defined later) of the current solution, estimate the accuracy of the neighbors, and select the neighbor with highest estimated accuracy. The algorithm terminates when we find a local optima. Empirical studies show that the accuracy of the \textit{LS} algorithm is significantly better than the \textit{2PG} algorithm, while the two algorithms are close to each other in terms of running time.

In the case of both algorithms, we consider several design choices and compare the accuracy of the variants of the algorithms that arise from different design choices. In the case of \textit{OrG} algorithm, the design choice is the sorting criteria used to sort the robots. For the \textit{LS} algorithm, design choices include the initial solution and the attack model.  

In summary, we make the following contributions:
\begin{itemize}
    \item 
    We propose two algorithms for the \textit{RCM} problem, namely, the \textit{OrG} algorithm and the \textit{LS} algorithm, which perform better than the state-of-the-art \textit{2PG} algorithm in terms of accuracy and running time. 
    \item
    We conduct extensive experiments with synthetic datasets to evaluate the accuracy and running time of our proposed algorithms with respect to the \textit{2PG} algorithm and a brute force optimal algorithm. 
    \item
    We perform a case study on the persistent monitoring problem to demonstrate the effectiveness of our proposed algorithms in a realistic application scenario.
\end{itemize}
    

\section{Problem Formulation}
\label{pf}


\subsection{Framework} 
\label{fw}

We adopt the framework introduced by Zhou et al.~\cite{lz} for the resilient multi-robot coverage problem. According to the proposed framework, there is a set of mobile robots, $\mathcal{R}$, which aim to cover a set of targets, $\mathcal{T}$. The targets can be mobile or stationary, distinguishable or indistinguishable, and can have a known or unknown motion model. It is assumed that the robots have perfect localization and can communicate with each other at all times. Using sensors, communication, and filtering techniques, the robots are able to calculate the estimated position of the stationary or moving targets as described in the framework proposed in~\cite{lz}. 

Time is divided into rounds of finite duration. We consider each round independently. At the beginning of a round, each robot generates a set of candidate trajectories, one of which will be followed in the current round. The set of trajectories of a robot $r$ is denoted be $\mathcal{P}_r$. The set of all robots' trajectories is denoted by $\mathcal{P}_\mathcal{R}$, i.e., $\mathcal{P}_\mathcal{R} := \cup_{r \in \mathcal{R}} \mathcal{P}_r$. Let $P$ be the number of all trajectories, i.e., $P = |\mathcal{P}_\mathcal{R}|$. Here the notation $|\mathcal{A}|$ denotes the cardinality of set $\mathcal{A}$.

\textbf{Target Coverage function:} The \textit{coverage} of a trajectory $p$ is defined as the set of targets that $p$ covers, which we denote by ${\rm C}(p)$. The \textit{target coverage function}, ${\rm F}$, takes as input a set of trajectories $\mathcal{P}$ and returns the number of unique targets covered by the trajectories in $\mathcal{P}$, i.e., ${\rm F}(\mathcal{P}) := |\bigcup_{p \in \mathcal{P}} {\rm C}(p)|$. 

Note that, the above definition of the target coverage function accounts for unweighted targets. If the targets are weighted, the target coverage function computes the sum of the weights of the targets covered by the trajectories in $\mathcal{P}$, i.e., ${\rm F}(\mathcal{P}) := \sum_{t \, \in \, \bigcup_{p \in \mathcal{P}} {\rm C}(p)} w(t)$, where $w(t)$ denotes the weight of target $t$. Our proposed algorithms can handle applications having weighted targets with no modification as we demonstrate in the persistent monitoring case study.

In both unweighted and weighted cases, the target coverage function F is \textit{monotone} and \textit{submodular}~\cite{rf24}. Other examples of monotone and submodular target coverage functions are mutual information and entropy~\cite{rf26}.

\textbf{Attack Model:} Throughout this paper, we use the words failure and attack interchangeably. We assume that at most $\alpha$ robots can fail (or, get attacked) at a time. We consider an optimal (i.e., worst-case) attack model as defined below. Given a set of trajectories $\mathcal{P}$, the target coverage function F, and an integer $\alpha$ denoting the maximum attack size, an \textit{optimal attack} on $\mathcal{P}$ of size $\alpha$ is defined as follows. 

\vspace{-0.2cm}

\[{\rm A}^*_\alpha(\mathcal{P}) := \argmin_{\mathcal{A} \subseteq \mathcal{P}} \hspace{0.1cm} {\rm F}(\mathcal{P} \backslash \mathcal{A}) \hspace{0.5cm} s.t. \hspace{0.1cm} |\mathcal{A}| \leq \alpha \] 

In other words, an optimal attack on $\mathcal{P}$ of size $\alpha$ is a subset of $\mathcal{P}$ of size at most $\alpha$ such that removal of the subset from $\mathcal{P}$ results in maximum decrease of the target coverage. In the above definition, the notation $\mathcal{A} \backslash \mathcal{B}$ denotes the set of elements in $\mathcal{A}$ that are not in $\mathcal{B}$. 

\subsection{Problem Definition} 
\label{pdef}

Given a set of targets, a set of robots $\mathcal{R}$, the trajectories for the robots $\mathcal{P}_\mathcal{R}$, the attack size $\alpha$, and a target coverage function F, the \textit{Resilient Coverage Maximization} (\textit{RCM}) problem aims to select a set of trajectories according to the following objective function.

\vspace{-0.3cm}

\begin{equation}
\label{pd}
\argmax_{\mathcal{S} \subseteq \mathcal{P}_\mathcal{R}} \hspace{0.1cm} {\rm F}(\mathcal{S} \backslash {\rm A}^*_{\alpha}(\mathcal{S})) \hspace{0.4cm} s.t. \hspace{0.1cm} |\mathcal{S} \cap \mathcal{P}_r|=1, \hspace{0.1cm} \forall r \in \mathcal{R} 
\end{equation}

In other words, the solution subset contains one trajectory per robot (enforced by the constraints), such that in the case of an optimal attack of size $\alpha$, the target coverage of the remaining robots is maximized. 

\subsection{Supplementary Definitions}
\label{sdef}

A \textit{Feasible Solution} is a subset of $\mathcal{P}_\mathcal{R}$ that satisfies the constraints in (\ref{pd}). In other words, a feasible solution corresponds to a valid assignment of trajectories to robots, i.e., one trajectory per robot.  The feasible solution that maximizes the objective function (\ref{pd}) is called the \textit{Optimal Solution}. We denote the optimal solution by $\mathcal{S^*}$. The \textit{Residual Coverage} of a feasible solution $\mathcal{S}$ is the number of targets covered by $\mathcal{S}$ after the optimal attack set is removed from $\mathcal{S}$. The residual coverage of $\mathcal{S}$ is denoted by ${\rm R}(\mathcal{S})$. According to the above definition, ${\rm R}(\mathcal{S}) = {\rm F}(\mathcal{S} \backslash {\rm A}^*_{\alpha}(\mathcal{S}))$.
\section{Ordered Greedy Algorithm}
\label{org}

In this section, we present a greedy algorithm for the \textit{RCM} problem that require O($P$) evaluations of the target coverage function F. The algorithm is named \textit{Ordered Greedy Algorithm} (\textit{OrG}) and is presented below (Algorithm 1). In this algorithm, first we sort the robots according to some \textit{sorting criteria} (Line 1). Then, for each robot (according to the sorted order), we greedily select the trajectory that maximizes the marginal coverage of the targets (Line 4-5).

\begin{algorithm}
    \caption{Ordered Greedy Algorithm}
    \textbf{Input:} $\mathcal{R}$, $\mathcal{P}_\mathcal{R}$, $\alpha$, F \\
    \textbf{Output:} Set of trajectories, $\mathcal{S}$ 
    \begin{algorithmic}[1]
        \STATE $<r_1, r_2, \ldots, r_{|\mathcal{R}|} > \, \leftarrow \, {\rm sort}(\mathcal{P}_\mathcal{R}, {\rm F}) $
        \STATE $\mathcal{S} \, \leftarrow \, \emptyset $
        \FOR{$i \, \leftarrow 1 \, {\rm to} \, |\mathcal{R}|$}
            \STATE $p^* \, \leftarrow \, \argmax_{p \in \mathcal{P}_{r_i}}{{\rm F}(\mathcal{S} \cup p)}$
            \STATE $\mathcal{S} \, \leftarrow \, \mathcal{S} \cup p^*$
        \ENDFOR
        \RETURN $\mathcal{S}$
  \end{algorithmic}
\end{algorithm}

To perform the sorting of the robots, for each robot $r$, we calculate a numerical value ${\rm V}(r)$, according to some sorting criteria, and then sort the robots in increasing or decreasing order of the assigned numerical value. Each criteria results in two variants of the \textit{OrG} algorithm: one for increasing order, and one for decreasing order. We use the following metrics as the sorting criteria.

\begin{itemize}
    \item 
    \textit{Size of Union of Target Coverage}: The numerical value of robot $r$ is the number of unique targets covered by all the trajectories of $r$, i.e., ${\rm V}(r) = |\cup_{p \in \mathcal{P}_r} {\rm C}(p)|$. The resultant \textit{OrG} algorithms are named \textit{OrG-U-I} and \textit{OrG-U-D} (for increasing and decreasing sorting order, respectively). 
    
    \item 
    \textit{Maximum Individual Target Coverage}: The numerical value of robot $r$ is the cardinality of the trajectory of $r$ that covers the maximum number of targets, i.e., ${\rm V}(r) = {\rm max}_{p \in \mathcal{P}_r} |{\rm C}(p)|$. The resultant algorithms are named \textit{OrG-M-I} and \textit{OrG-M-D}.
    
\end{itemize}

We also consider another variant where the ordering of the robots is random (\textit{OrG-R}). Note that, each of the above algorithms requires O($P$) evaluations of F. Here, $P$ is the sum of number of candidate trajectories of all robots. To calculate the numerical values of the robots, we need $P$ evaluations of F. Also, in Line 4 of Algorithm 1, the call to F is executed $P$ times in total. Thus, the total number of evaluations of F for the \textit{OrG} algorithm is O($P$).
\section{Local Search Algorithm}
\label{ls}

In this section, we describe an algorithm based on the local search technique. In a traditional local search algorithm, we start with an initial solution. In each iteration, we make small local changes to the current solution to form a set of \textit{neighbor} solutions. Then we evaluate the objective function on the neighbors to determine if any improvement over the current solution is possible. If a better solution is found, the search moves to that direction. Otherwise, the algorithm terminates. 

A tricky aspect of adopting a local search based approach to the \textit{RCM} problem is that evaluating the objective function for a given solution is not straightforward. In this problem, the objective value of a feasible solution $\mathcal{S}$ is ${\rm F}(\mathcal{S} \backslash {\rm A}^*_\alpha(\mathcal{S}))$. Thus, given a feasible solution $\mathcal{S}$, in order to evaluate the objective function for $\mathcal{S}$, we need to construct an optimal attack on $\mathcal{S}$. However, constructing an optimal attack on $\mathcal{S}$ is an NP-hard problem, because the \textit{Maximum k-Coverage Problem}, which is known to be NP-hard, can be reduced to the problem of constructing an optimal attack~\cite{nphard}. Consequently, in this algorithm, we use computationally feasible greedy attack models, instead of an optimal attack model, to drive the local search. We denote the greedy attack function by A, which is further discussed later in this section. 

\begin{algorithm}
    \caption{Local Search Algorithm}
    \label{algols}
    \textbf{Input:} $\mathcal{R}$, $\mathcal{P}_\mathcal{R}$, $\alpha$, F \\
    \textbf{Output:} Set of trajectories, $\mathcal{S}$ 
    \begin{algorithmic}[1]
        \STATE $\mathcal{S} \, \leftarrow \, {\rm INIT}(\mathcal{P}_\mathcal{R}, {\rm F})$
        \STATE $z \, \leftarrow \, {\rm F}(\mathcal{S} \backslash {\rm A}_\alpha(\mathcal{S}))$
        \WHILE{TRUE}
            \STATE $f \, \leftarrow \, {\rm FALSE}$
            \FOR{all neighbor $\dot{\mathcal{S}}$ of $\mathcal{S}$}
                \STATE $\mathcal{A} \, \leftarrow \,{\rm A}_\alpha(\dot{\mathcal{S}})$
                \STATE $\dot{z} \, \leftarrow \, {\rm F}(\dot{\mathcal{S}} \backslash \mathcal{A})$
                \IF{$\dot{z} > z$}
                    \STATE $f, \mathcal{S}, z \, \leftarrow \,{\rm TRUE}, \dot{\mathcal{S}}, \dot{z}$
                    \STATE \textbf{break}
                \ENDIF
            \ENDFOR
            \IF{$f = {\rm FALSE}$}
                \STATE \textbf{break}
            \ENDIF
        \ENDWHILE
        \RETURN $\mathcal{S}$
  \end{algorithmic}
\end{algorithm}

Now we describe the local search algorithm (Algorithm 2) in detail. We start with an initial feasible solution $\mathcal{S}$ (Line 1) and the corresponding objective value $z$ (Line 2). In each iteration of the local search (Line 3--16), we consider all neighbors of the current solution $\mathcal{S}$ (Line 5). Any feasible solution which differs from $\mathcal{S}$ by exactly one trajectory is defined to be a \textit{neighbor} of $\mathcal{S}$. For each neighbor $\dot{\mathcal{S}}$ of $\mathcal{S}$, we construct a greedy attack on $\dot{\mathcal{S}}$, denoted by $\mathcal{A}$ (Line 6), and calculate the corresponding objective value $\dot{z}$ (Line 7). If the neighbor solution is better than the current solution (Line 8), we restart the iteration with updated solution and objective value (Line 9--10). If no neighbor leads to a solution better than the current solution, the local search terminates (Line 4, 13--14), and the current solution is returned (Line 17) as the solution of the \textit{RCM} problem. 

Note that, $\dot{z}$ (computed in Line 7) is rather an estimation of the residual coverage of $\dot{\mathcal{S}}$, because we use a non-optimal greedy attack model ${\rm A}$ instead of the optimal attack model ${\rm A}^*$ to compute the residual coverage. We use A instead of ${\rm A}^*$ so that the \textit{LS} algorithm runs in polynomial time.

Several variants of the \textit{LS} algorithm arise when we use different attack models (Line 2, 6) and different initial solutions (Line 1). We consider the following two greedy attack models. In both models, the attacker takes as input a feasible solution $\mathcal{S}$, and returns a subset of $\mathcal{S}$ of size $\alpha$. 

\begin{itemize}
    \item
    Attack Model 1 (A1): We initialize a set $\mathcal{X}$ as empty set. In each iteration, we determine the trajectory in $\mathcal{S} \backslash \mathcal{X}$ addition of which maximizes the marginal \textit{increase} in target coverage of $\mathcal{X}$, and add that trajectory to $\mathcal{X}$. After $\alpha$ iterations, we return $\mathcal{X}$. Note that, A1 seeks to maximize the coverage of $\mathcal{X}$, and the coverage of $\mathcal{X}$ selected by A1 is within a factor of $1 - \frac{1}{e}$ of the optimal~\cite{e11}.
    
    \item
    Attack Model 2 (A2): We initialize a set $\mathcal{X}$ to $\mathcal{S}$. In each iteration, we determine the trajectory in $\mathcal{X}$ removal of which maximizes the marginal \textit{decrease} in target coverage of $\mathcal{X}$, and remove that trajectory from $\mathcal{X}$. After $\alpha$ iterations, we return $\mathcal{S} \backslash \mathcal{X}$.
    
\end{itemize}

We consider two initial solutions as follows. 

\begin{itemize}
    \item
    Initial Solution 1 (I1): The output of the Oblivious Greedy algorithm (to be described in Section~\ref{expsetup}).
    
    \item
    Initial Solution 2 (I2): The output of the \textit{OrG-U-I} algorithm. We choose \textit{OrG-U-I} because it empirically performs better than the other \textit{OrG} variants.
    
\end{itemize}

Two attack models and two initial solutions yield four variants of the \textit{LS} algorithm. We append the attack type and initial solution type to name the \textit{LS} variants. For example, \textit{LS-A1-I2} is the variant of the \textit{LS} algorithm that uses attack model A1 and initial solution I2.  

Note that, in the \textit{LS} algorithm, the number of evaluations of F is dominated by the number of calls to the attack function in Line 6. In the case of both of the above attack models, constructing a greedy attack requires O($\alpha R$) calls to F, where $R$ is the number of robots. There are O($P$) neighbors of the current solution $\mathcal{S}$. Consequently, for both attack models, the \textit{LS} algorithm requires O($I \alpha RP$) evaluations of F, where $I$ is the number of times the local search iterates. However, we reduce the running time of the \textit{LS} algorithm by a factor of O($R$) as described in the next section.

\subsection{Acceleration of the Local Search Algorithm}
\label{accds}

To accelerate the \textit{LS} algorithm, we  reduce the time required to compute the attack functions A1 and A2. First we analyze the time complexity of a straightforward way of computing the attack functions. Let $t^*$ denote the maximum number of targets covered by one trajectory. Recall from previous section that constructing greedy attack A1 or A2 requires O($\alpha R$) evaluations of target coverage function. The time complexity of calculating the target coverage of a set of trajectories, $\mathcal{P}$, is O($R t^*$), because, in our application, $|\mathcal{P}| \leq R$. Thus, a straightforward implementation of the attack function takes O($\alpha R^2 t^*$) time.   

Now we present a way to accelerate the process of computing the attack functions A1 and A2. The main idea is to store precalculated results in an auxiliary data structure, and thus avoid computing the attack function from scratch within each greedy iteration. Let $T$ denote the number of targets. We maintain an array $Y$ of size $T$, which stores, for each target $t$, how many trajectories in $\mathcal{X}$ covers $t$. We update the auxiliary array once in each greedy iteration when a trajectory is added to $\mathcal{X}$ (A1) or deleted from $\mathcal{X}$ (A2), which takes O($t^*$) time. Note that, by using the array $Y$, we can determine the marginal increase or decrease in target coverage of $\mathcal{X}$, when a trajectory is added to or removed from $\mathcal{X}$, in O($t^*$) time. Consequently, each greedy iteration of the attack functions takes O($Rt^*$) time. This makes the overall computation time of the attack functions O($\alpha R t^*$). Thus, using the above described acceleration technique, we achieve a performance speedup of a factor of O($R$) over the straightforward implementation.

\section{Experiments}
\label{exp}

In this section, we empirically evaluate our proposed algorithms and present the experimental results. First, we discuss the experimental setup (Section~\ref{expsetup}). Next, we compare the accuracy (Section~\ref{comp_a}) and running time (Section~\ref{comp_et}) of our proposed algorithms with the \textit{2PG} algorithm. We present the results of a sensitivity analysis in Section~\ref{appsens}. Finally, we list the key findings of the experiments (Section~\ref{findings}).  

\subsection{Experimental Setup}
\label{expsetup}

\textbf{Evaluation Metric}: We use two metrics to empirically evaluate our proposed algorithms: accuracy and running time. The \textit{accuracy} of a feasible solution $\mathcal{S}$ is the ratio of the residual coverages of $\mathcal{S}$ and $\mathcal{S^*}$, where $\mathcal{S^*}$ is the optimal solution. Thus, the accuracy of a feasible solution $\mathcal{S}$ is a measure of the quality of $\mathcal{S}$ with respect to the optimal solution $\mathcal{S^*}$. If, in an experiment, the optimal solution is known, we directly report the accuracy (with respect to the optimal solution) of the solutions found by the algorithms which are being compared. On the other hand, if the optimal solution is unknown, we compute the residual coverages of the solutions found by the algorithms and report the relative accuracy with respect to the \textit{2PG} algorithm. Note that, a higher residual coverage corresponds to higher accuracy, and vice versa, since the ratio of residual coverages of two feasible solutions equals the ratio of their accuracy.

\begin{figure*}[!ht]
\centering
\includegraphics[width=1.00\linewidth]{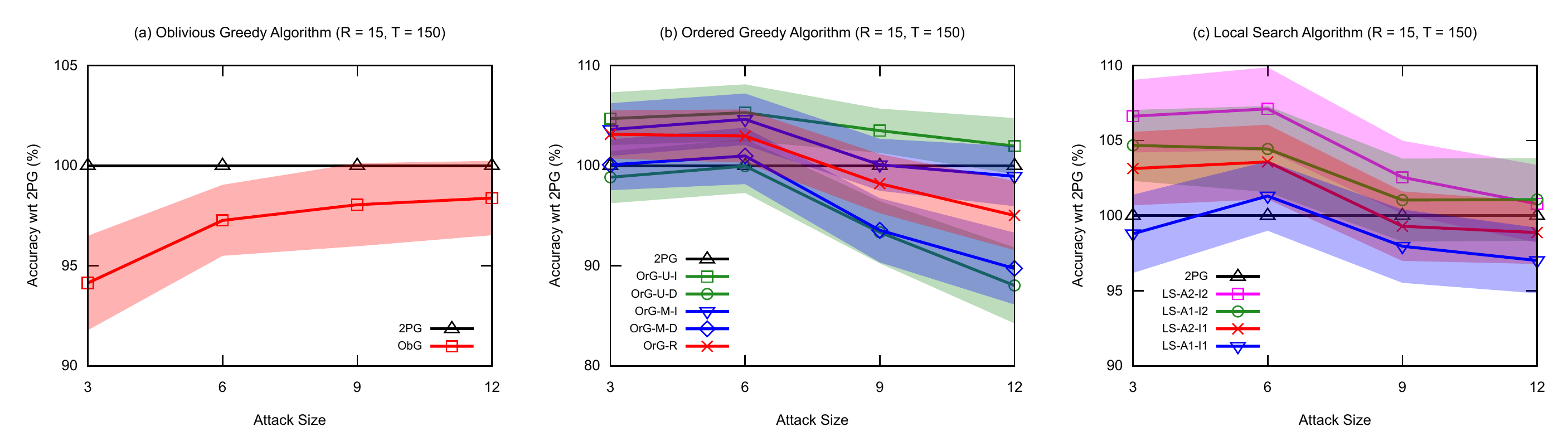}
\caption{Comparison of accuracy of (a) \textit{ObG} algorithm, (b) \textit{OrG} algorithm, and (c) \textit{LS} algorithm with \textit{2PG} algorithm.}
\label{f1g}
\end{figure*}


\textbf{Compared Algorithms}: We empirically compare the performance of our proposed algorithms (\textit{OrG} and \textit{LS}) with the \textit{2PG} algorithm. We additionally consider two baseline algorithms as follows:
\begin{itemize}
    \item 
    \textit{Brute Force algorithm}: The Brute Force (\textit{BF}) algorithm determines the optimal solution of the \textit{RCM} problem. In the \textit{BF} algorithm, we formulate the \textit{RCM} problem as an Integer Linear Program (ILP), and use a commercial MILP solver (Gurobi~\cite{grb}) to solve the ILP. The ILP formulation of the \textit{RCM} problem is given in Appendix~\ref{ilp}. Note that, this ILP formulation has an exponential number of constraints with respect to the number of robots. Hence, it can be used to solve only very small instances of the \textit{RCM} problem.
    
    \item 
    \textit{Oblivious Greedy algorithm}: In the Oblivious Greedy (\textit{ObG}) algorithm, we select, for each robot, the trajectory that covers maximum number of targets. Formally, the solution found by this algorithm is $\bigcup_{r \in \mathcal{R}} \argmax_{p \in \mathcal{P}_r} {\rm F} (\{ p \})$. The \textit{ObG} algorithm makes $P$ calls to the target coverage function F. 
    
\end{itemize}

\begin{figure}[!ht]
\centering
\includegraphics[width=0.50\linewidth]{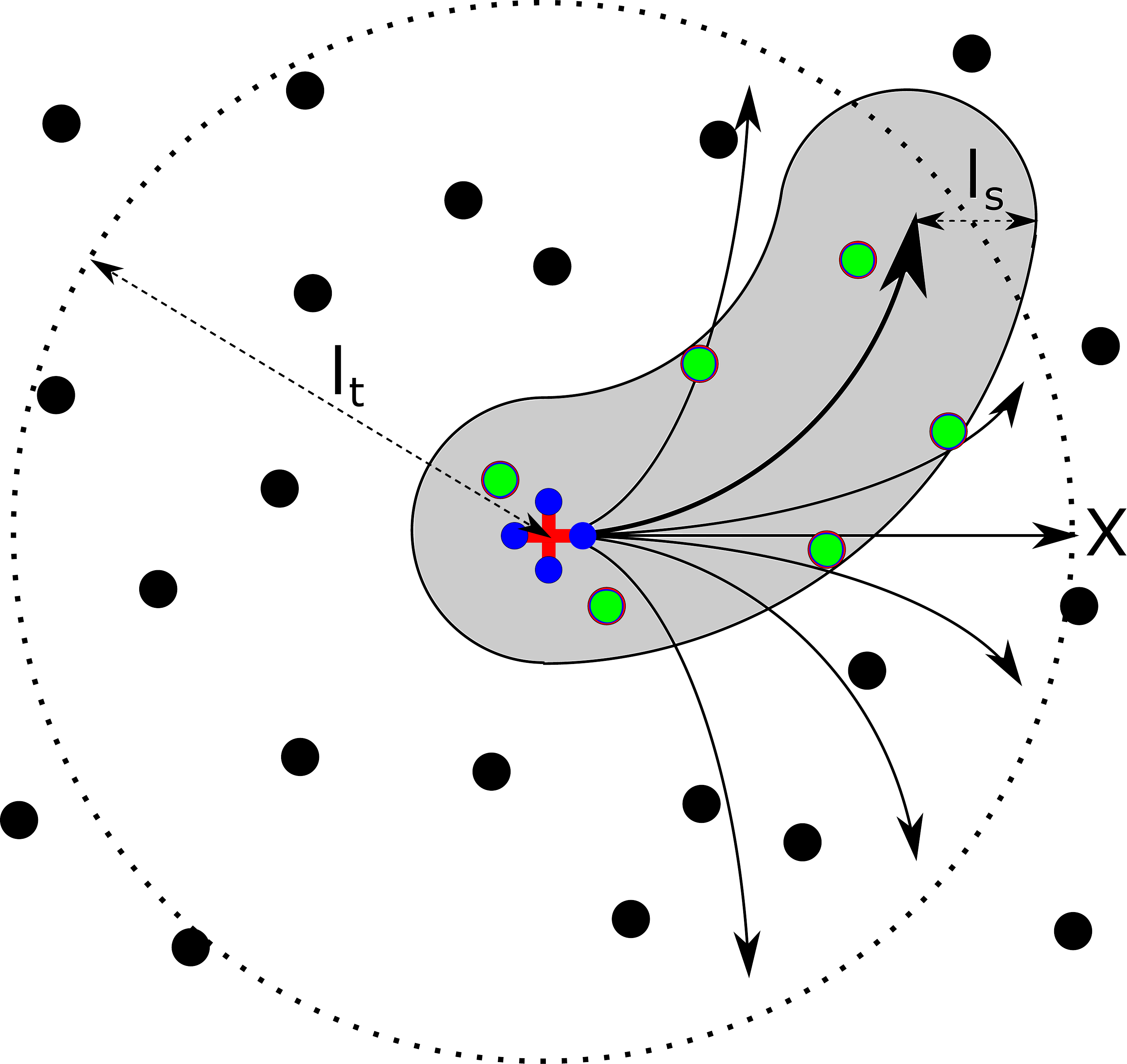}
\caption{Dataset and trajectory generation.}
\label{expfig}
\end{figure}

\textbf{Dataset}: In our experiments, we use a synthetic dataset generated as follows. First, we select the locations of the targets and robots within a $100 \times 100 \, m^2$ 2D region with uniform probability. For each robot, we consider 7 elliptical candidate trajectories each of length $l_t$ as shown in Figure~\ref{expfig}. The candidate trajectories are centered around the current direction of the robot (towards X in Figure~\ref{expfig}). A trajectory $\tau$ covers all the targets located within a distance of $l_s$ from $\tau$. In Figure~\ref{expfig}, we show the coverage region of the bold trajectory in grey, and the covered targets in green. The trajectory generation procedure described above is suitable for kinodynamic planning and  commonly used in standard literature~\cite{rf11}. The procedure imitates a scenario in which a set of targets on the ground are being covered by a set of UAVs with downward facing cameras.

\textbf{Parameter Set}: We use different sets of parameter values for different experiments. For example, in the experiments where we compute a brute force solution, we use only 6 robots to keep the total running time low. For other experiments, we use higher number of robots. Also, we set $l_0$ and $l_t$ according to the number of robots to ensure that there is sufficient overlap among the trajectories. The values of the parameters used in each experiment is mentioned in the corresponding subsection. Each experiment is conducted 100 times and the average is reported. We assume that the number of robot failures is equal to the attack size, $\alpha$.

\textbf{Platform}: The algorithms are implemented using C++. The experiments are conducted on a core-i7 2GHz PC with 8GB RAM, running Microsoft Windows 10.

\subsection{Comparison of Accuracy}
\label{comp_a}

\textbf{Relative Accuracy with respect to \textit{2PG} Algorithm}: In this experiment, we construct a dataset with 15 robots, 150 targets, $l_t=40m$, and $l_s=10m$. We vary the attack size in increments of 3 and report the average relative accuracy (in percentage) of our proposed algorithms with respect to the \textit{2PG} algorithm under an optimal attack model (Figure~\ref{f1g}). The standard deviation is shown using the shades.

The experimental results show that the accuracy of the \textit{ObG} algorithm is consistently lower than the \textit{2PG} algorithm (Figure~\ref{f1g}(a)). In the case of \textit{OrG} algorithm, the \textit{OrG-I} variants have higher accuracy than their \textit{OrG-D} counterparts (Figure~\ref{f1g}(b)). The accuracy of the \textit{2PG} and \textit{OrG-R} algorithms lie in between the \textit{OrG-I} and \textit{OrG-D} variants. 

We claim that the increasing sorting order leads to an even distribution of the targets to trajectories. Consequently, the reduction of target coverage after an optimal attack is smaller in the case of \textit{OrG-I} variants as opposed to the \textit{OrG-D} ones. We empirically verify the correctness of the above claim by conducting an experiment where we compare the standard deviations of the marginal coverages in each greedy iteration of \textit{OrG-U-I} and \textit{OrG-U-D}. We find that the standard deviation of the \textit{OrG-D} variant is on average $60\%$ higher than the \textit{OrG-I} variant, which provides empirical evidence in support of our claim. We do not show the experimental results in detail for brevity of presentation.

In the case of \textit{LS} algorithm, experimental results depicted in Figure~\ref{f1g}(c) show that attack model 2 (A2) leads to better accuracy than attack model 1 (A1). Also, initial condition 2 (I2) gives higher accuracy in comparison to initial condition 1 (I1). Thus, \textit{LS-A2-I2} has the highest accuracy among the \textit{LS} variants. Also, the accuracy of \textit{LS-A2-I2} is significantly higher than the \textit{2PG} algorithm across all attack sizes.

In the above experiments, we observe that in the case of OrG and LS algorithms, the accuracy decreases as we increase $\alpha$. We conjecture that when $\alpha$ is large, the room for optimization is limited. In other cases, there is more room for optimization, and consequently, our proposed algorithms perform relatively better.

For brevity of presentation, from now on, instead of reporting the accuracy of all the variants of our proposed algorithms, we only report the results for the \textit{OrG} and \textit{LS} variants with highest accuracy, namely, \textit{OrG-U-I} and \textit{LS-A2-I2} respectively, along with the \textit{ObG} and \textit{2PG} algorithm.



\textbf{Accuracy with respect to Brute Force Algorithm}: In this experiment, we determine the accuracy of our proposed algorithms. Note that, the accuracy of a feasible solution $\mathcal{S}$ is the ratio of the residual coverages of $\mathcal{S}$ and $\mathcal{S^*}$, the optimal solution. We compute the optimal solution using the \textit{BF} algorithm, which requires very high computational time. Consequently, in this experiment, we consider small instances of the \textit{RCM} problem with 6 robots, 60 targets, and $l_t=50m$ and $l_s=15m$, and use attack sizes 2, 3, and 4.

\begin{figure}[!ht]
\centering
\includegraphics[width=0.70\linewidth]{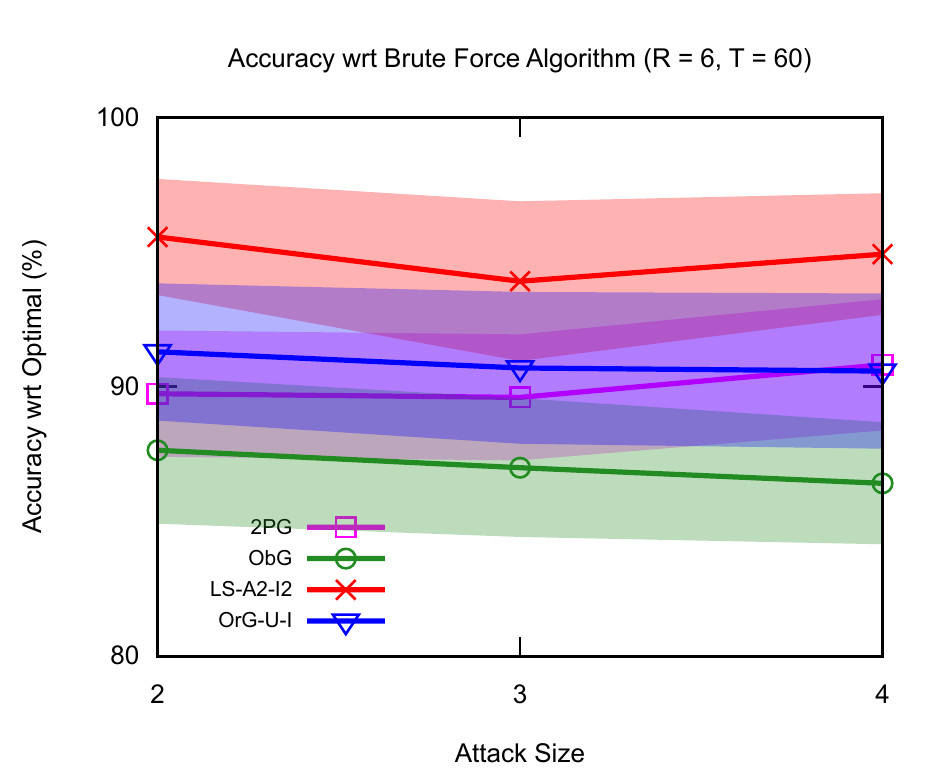}
\caption{Comparison of accuracy of proposed algorithms with \textit{BF} algorithm.}
\label{f2g}
\end{figure}

The experimental results in Figure~\ref{f2g} show that the \textit{LS} algorithm has the highest accuracy, followed by \textit{OrG}, \textit{2PG}, and \textit{ObG} algorithms in the above order, which is in accordance with the experimental results presented in the previous section. Note that, the accuracy of the \textit{BF} algorithm is 100\%.

\begin{figure}[!ht]
\centering
\includegraphics[width=0.70\linewidth]{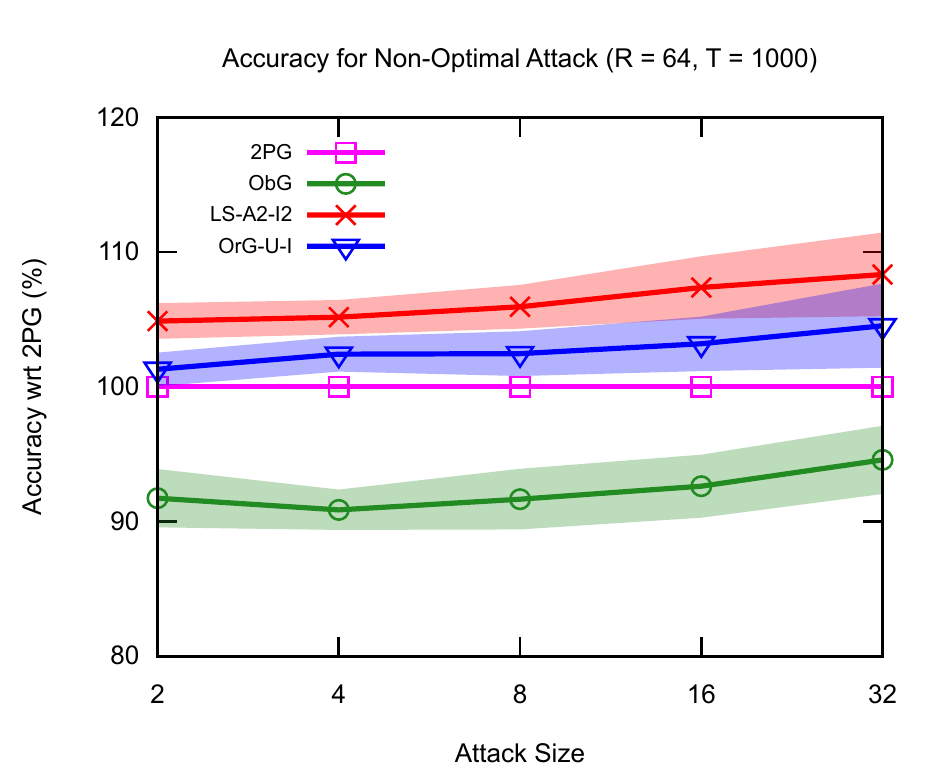}
\caption{Comparison of accuracy with Non-optimal Attack Model.}
\label{f3g}
\end{figure}

\textbf{Relative Accuracy with respect to \textit{2PG} Algorithm for Large Problem Instances:} In this experiment, we evaluate the accuracy of our proposed algorithms for large instances of the \textit{RCM} problem. In the case of large problem instances, it is not feasible to compute the residual coverage, because constructing an optimal attack requires exponential time with respect to the number of robots. Consequently, we resort to a non-optimal greedy attack model to compute an estimation of the residual coverage. We use attack model 2 (A2), outlined in Section~\ref{ls}, which is a greedy approximate attack model computable in polynomial time.

In this experiment, we use 64 robots, 1000 targets, $l_t=25m$, and $l_s=5m$, and vary the attack size in factors of 2, and report the relative accuracy with respect to the \textit{2PG} algorithm. The experimental results (Figure~\ref{f3g}) show that the accuracy of the proposed algorithms with attack model A2 is equivalent to the accuracy found in previous sections using an optimal attack model. The \textit{LS} algorithm still has the best accuracy among the compared algorithms.

\subsection{Comparison of Running Time}
\label{comp_et}

\begin{figure}[!ht]
\centering
\includegraphics[width=0.70\linewidth]{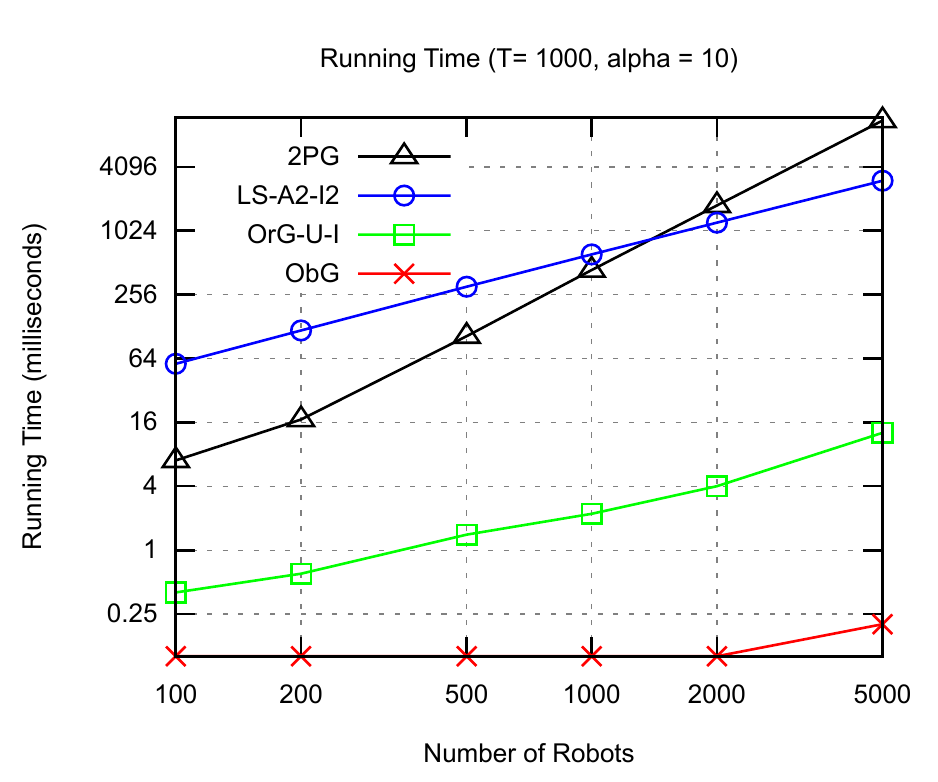}
\caption{Comparison of running time.}
\label{fet}
\end{figure}

In this experiment, we vary the number of robots from 100 to 5000, and use 1000 targets and an attack size of 10. The experimental results (Figure~\ref{fet}) show that the \textit{ObG} algorithm has the lowest running time, followed by the \textit{OrG} algorithm. The \textit{LS} algorithm and the \textit{2PG} algorithm run slower than the former two algorithms, and the \textit{LS} algorithm outperforms the \textit{2PG} algorithm as the number of robots goes past 1000. The experimental results are in accordance with the time complexity analysis presented in the previous sections. Through another set of experiments (not presented in the paper for brevity), we find that other variants of the \textit{OrG} and \textit{LS} algorithms have similar running time as the counterpart compared above. We also find that, increasing the number of candidate trajectories increases the running time linearly, and increasing the number of targets increases the running time slowly.

\subsection{Sensitivity Analysis}
\label{appsens}

\begin{figure}[!h]
\centering
\includegraphics[width=0.70\linewidth]{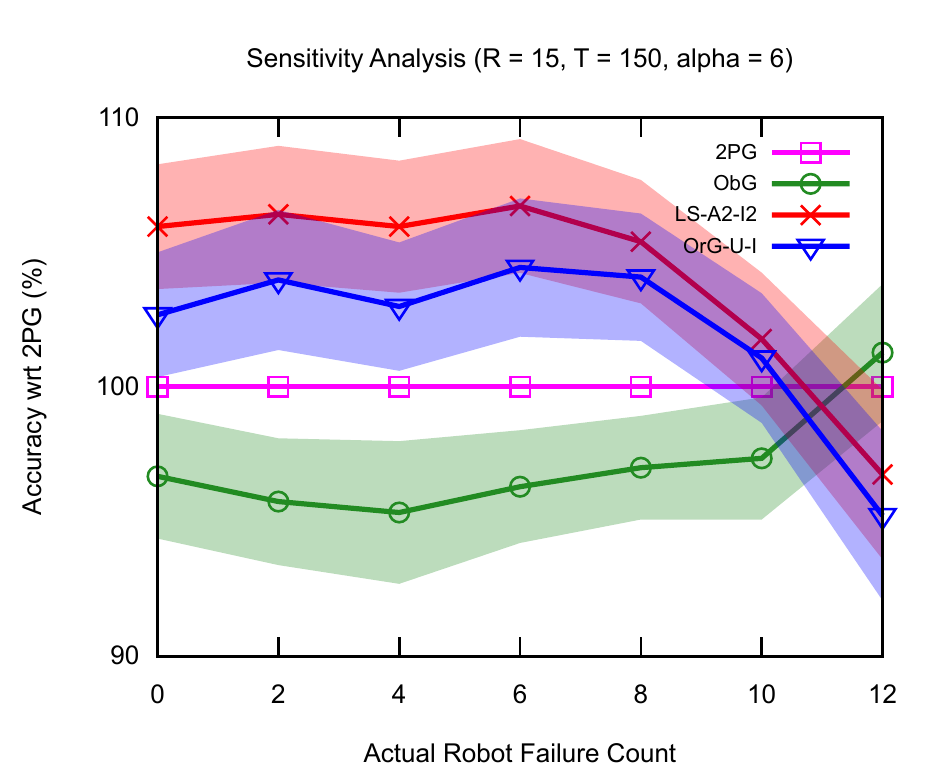}
\caption{Sensitivity analysis.}
\label{f4g}
\end{figure}

In the above experiments, we have assumed that the actual number of robot failures equals the attack size, $\alpha$. However, in reality, the actual number of robot failures may be more or less than $\alpha$. For example, in a practical deployment of 10 robots which assumes $\alpha=2$, there may be no robot failure. Hence, we present experimental results where the number of robot failures differs from the attack size.

In this experiment, we consider a scenario with 15 robots, 150 targets, and attack size, $\alpha=6$. We vary the number of robot failures (worst-case failure) in increments of 2 and report the relative accuracy of our proposed algorithms with respect to the \textit{2PG} algorithm. 

The experimental results show that, the accuracy of the proposed algorithms (\textit{OrG-U-I} and \textit{LS-A2-I2}) drops sharply if the number of robot failure exceeds $\alpha$. If the number of robot failures is within the assumed maximum attack size (i.e., less than or equal to $\alpha$), the proposed algorithms give better accuracy than the \textit{2PG} algorithm.

\subsection{Key Findings}
\label{findings}

The key findings of the experiments are listed below:

\begin{itemize}
    \item 
    In the case of \textit{OrG} algorithm, \textit{OrG-I} variants show higher accuracy than \textit{OrG-D} variants. In the case of \textit{LS} algorithm, \textit{LS-A2} variants have higher accuracy than \textit{LS-A1} variants. \textit{OrG-U-I} and \textit{LS-A2-I2} have the highest accuracy within their respective class. Both \textit{LS-A2-I2} and \textit{OrG-U-I} have higher accuracy than \textit{2PG} algorithm, with \textit{LS-A2-I2} slightly outperforming \textit{OrG-U-I} (Section~\ref{comp_a}). 
    \item
    The compared algorithms exhibit similar empirical performances, when evaluated using attack model A2 and an optimal attack model. This result advises the use of computationally light attack model A2 in the case of large problem instances (Section~\ref{comp_a}).
    \item
    \textit{ObG} and \textit{OrG} algorithms run orders of magnitude faster than \textit{2PG} and \textit{LS} algorithms, while \textit{2PG} and \textit{LS} algorithms have comparable running time. The empirical running times are in accordance with the theoretical time complexity analysis (Section~\ref{comp_et}).
    \item
    If the number of actual robot failures is less than $\alpha$, the performance of the proposed algorithms do not suffer. This result suggests that when the true attack size is unknown, it is better to overestimate $\alpha$ than underestimating it (Section~\ref{appsens}).
    
\end{itemize}

\section{Case Study: Resilient Persistent Monitoring}
\label{permon}

In this section, we evaluate our proposed algorithms in the context of a practical application scenario, i.e., the \textit{Resilient Persistent Monitoring} (\textit{RPM}) problem. In a typical setup of the persistent monitoring problem, multiple robots monitor a 2D grid-based environment with obstacles. Each non-obstructed grid-cell $c$ in the environment has a latency value (denoted by $l_c$) in the range $[0, l_{max}]$. The latency of a cell vary according to the last time the cell was visible from some robot. If a cell $c$ is visible from some robot in the current time step, $l_c$ is set to $0$. Otherwise, if $c$ is visible from no robots in the current time step, $l_c$ increases linearly at each time step, until it reaches $l_{max}$. In the traditional persistent monitoring problem, for each robot, we select one trajectory from a set of candidate trajectories, such that, the overall decrease in latency (i.e., sum of reduction in latency values of all cells) is maximized. In the \textit{RPM} problem, we select the trajectories such that in the case of a worst-case failure of at most $\alpha$ robots, the overall decrease in latency achieved by the rest of the robots is maximized.

\begin{figure}[!ht]
\centering
\includegraphics[width=1.0\linewidth]{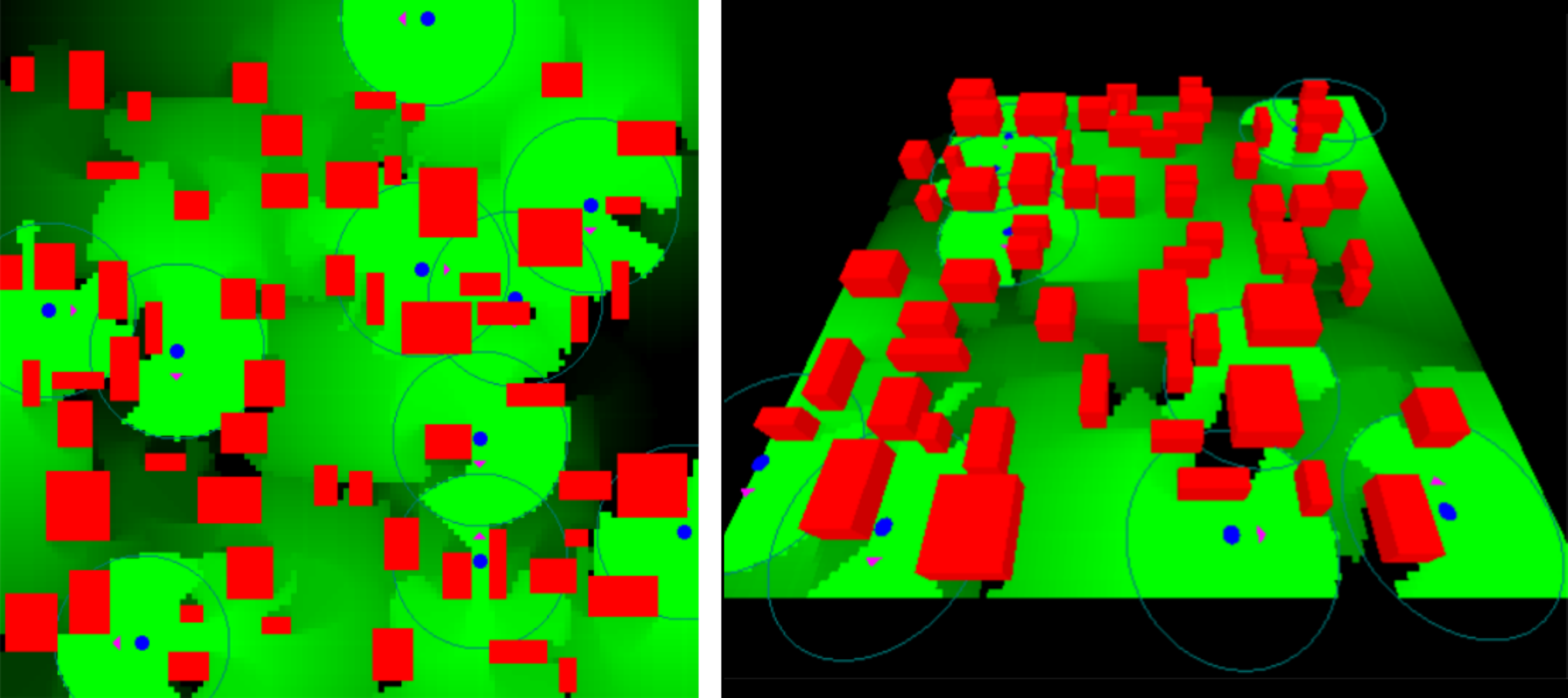}
\caption{Top down and perspective views are shown in left and right respectively.  Blue circles and red boxes represent robots and obstacles respectively. Purple arrows show the direction of the current trajectory of robots. Shades between green and black represent latency of the cells, where green and black stands for $0$ and $l_{max}$ respectively. }
\label{snap}
\end{figure}

Note that, our proposed algorithms for the \textit{RCM} problem can be suitably modified to solve the \textit{RPM} problem. Essentially, \textit{RPM} problem is a weighted version of the \textit{RCM} problem, in that each cell serves as a target object and the weight of a cell $c$ is equal to the reduction in the latency of $c$ when $c$ becomes visible, i.e., $l_c$. Hence, in the \textit{RPM} problem, the target coverage function F computes the sum of the latency values of the cells covered by a set of trajectories. 

\begin{figure}[!ht]
\centering
\includegraphics[width=0.70\linewidth]{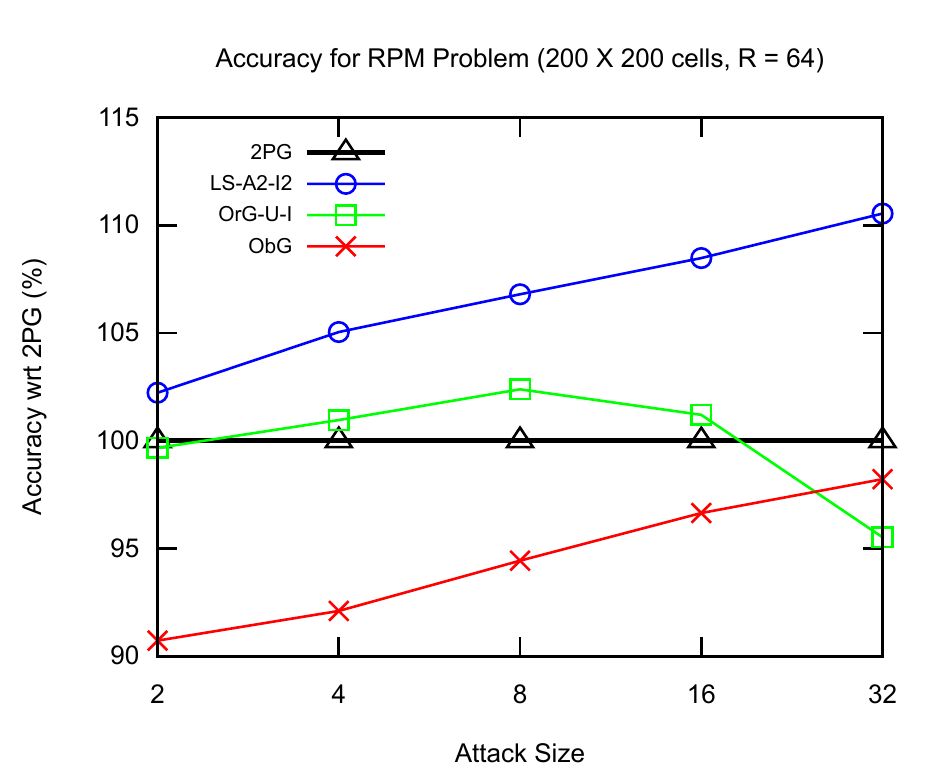}
\caption{Comparison of accuracy for \textit{RPM} problem.}
\label{rpm}
\end{figure}

We use a 2D environment having $200 \times 200$ cells, and 100 obstacles which occupy approximately 15\% of the environment. There are 64 robots, each with a $360^{\circ}$ view of the environment occluded by obstacles, and a visibility range of 15 times the length of a cell. Each robot has 4 linear candidate trajectories, $\{$forward, backward, left, right$\}$, one of which it will select. $l_{max}$ is set to 100 and the latency of non-visible cells are set to increase by 1 unit per time step. Two sample partial snapshots of the environment implemented using OpenGL\footnote{\url{https://youtu.be/XdQ5h5aOMAA}} are shown in Figure~\ref{snap}. 

The experimental results are presented in Figure~\ref{rpm}. Because the problem instance is fairly large, we use non-optimal attack model A2 instead of the optimal attack model, and use the same set of parameters as used in the last experiment of Section~\ref{comp_a}. Experimental results show that the performance of the algorithms is consistent with the findings of Section~\ref{exp}. 

\section{Conclusion}
\label{con}

In this work, we have proposed two algorithms for the coverage maximization problem with multiple robots in an adversarial setting. Our proposed algorithms have outperformed the state-of-the-art algorithm in terms of accuracy and running time. We have demonstrated the effectiveness of our proposed solutions by conducting empirical studies. 

In future, we intend to evaluate real-world deployment of our proposed algorithms in surveillance and patrolling. One may also consider reformulating the problem with a computationally feasible non-optimal attack model, and reevaluate the performance of the discussed algorithms.



\appendices

\section{ILP Formulation}
\label{ilp}

First we introduce some notations related to the ILP formulation of the \textit{RCM} problem. The set of trajectories that cover target $t$ is denoted by $\mathcal N_t$, i.e., $\mathcal N_t = \{p \in \mathcal P_{\mathcal R} : p \, \, {\rm covers} \, \, t \}$. The set of all subsets of $\mathcal R$ of size at most $\alpha$ is denoted by $\mathcal W$. Essentially, each element of $\mathcal W$ represents an attack of size at most $\alpha$. The set of all trajectories pertaining to a given a subset $R$ of the robots, i.e., $R \subseteq \mathcal R$, is denoted by $\mathcal V_R$, i.e., $\mathcal V_R = \cup_{r \in R} \mathcal P_r$. Now we introduce the variables used in the ILP formulation of the \textit{RCM} problem.

\begin{itemize}
    \item For each candidate trajectory $p \in \mathcal P_\mathcal R$, we use one binary variable $x_p$, which indicates if the candidate trajectory $p$ is selected.
    
    
    \item For each pair $(t,w)$ such that $t \in \mathcal T$ and $w \in \mathcal W$, we use one binary variable $y_{t,w}$, which indicates whether the target $t$ is covered when the robots in the set $w$ fail.
    
    \item One integer variable $z$ which indicates the maximum target coverage achieved under an optimal attack model.
    
\end{itemize}

The ILP formulation is presented below. The set of constraints in \ref{ilpc1} enforce that exactly one candidate trajectory is selected for each robot. Constraints \ref{ilpc2} ensure that, if a target $t$ is covered under attack $w$, i.e., $y_{t,w} = 1$, at least one candidate trajectory, which covers $t$ and which pertains to no robot in $w$, is selected. Constraints \ref{ilpc3} along with objective are used to maximize the post-attack coverage over all possible attacks. The binariness and integrality conditions of the variables are omitted for brevity of presentation.


$${\rm Maximize:} \; \; z $$


\begin{equation}
\label{ilpc1}
{\rm Subject \; to:} \; \; \sum_{p \in \mathcal P_r} \: x_p = 1 \hspace{1 cm} \forall r \in \mathcal R
\end{equation}  


\begin{equation}
\label{ilpc2}
\sum_{p \in \mathcal N_t \backslash \mathcal V_w} \: x_p \geq y_{t,w} \hspace{1 cm} \forall t \in \mathcal T, \, \, \forall w \in \mathcal W
\end{equation}


\begin{equation}
\label{ilpc3}
\sum_{t \in \mathcal T} \: y_{t,w} \geq z \hspace{1 cm} \forall w \in \mathcal W
\end{equation}


\begin{thebibliography}{00}
 \bibitem{lz} L. Zhou, V. Tzoumas, G. J. Pappas, and P. Tokekar, ``Resilient active target tracking with multiple robots,'' IEEE Robotics and Automation Letters, vol. 4, pp. 129--136, 2018.
 
 \bibitem{rf1} P. Tokekar, E. Branson, J. Vander Hook, and V. Isler, ``Tracking aquatic invaders: Autonomous robots for monitoring invasive fish,'' IEEE Robotics \& Automation Magazine, vol. 20, pp. 33--41, 2013.
 
 \bibitem{rf2} B. Grocholsky, J. Keller, V. Kumar, and G. Pappas, ``Cooperative air and ground surveillance,'' IEEE Robotics \& Automation Magazine, vol. 13, no. 3, pp. 16--25, 2006.
 
 
\bibitem{rf4} V. Kumar, and N. Michael, ``Opportunities and challenges with autonomous micro aerial vehicles,'' The International Journal of Robotics Research, vol. 31, no. 11, pp. 1279--1291, 2012.

\bibitem{rf5} N. Atanasov, J. Le Ny, K. Daniilidis, and G. J. Pappas, ``Information acquisition with sensing robots,'' in IEEE International Conference on Robotics and Automation, 2014, pp. 6447--6454.

\bibitem{rf6} C. Robin, and S. Lacroix, ``Multi-robot target detection and tracking: taxonomy and survey,'' Autonomous Robots, vol. 40, pp. 729--760, 2016.

\bibitem{rf7} J. R. Spletzer, and C. J. Taylor, ``Dynamic sensor planning and control for optimally tracking targets,'' The International Journal of Robotics Research, vol. 22, no. 1, pp. 7--20, 2003.



\bibitem{rf10} A. Pierson, Z. Wang, and M. Schwager, ``Intercepting rogue robots: An algorithm for capturing multiple evaders with multiple pursuers,'' IEEE Robotics and Automation Letters, vol. 2, pp. 530--537, 2017.

\bibitem{rf11} B. Schlotfeldt, N. Atanasov and G. J. Pappas, ``Maximum Information Bounds for Planning Active Sensing Trajectories,'' Intl. Conference on Intelligent Robots and Systems (IROS), 2019, pp. 4913--4920.


\bibitem{rf13} E. Sless, N. Agmon, and S. Kraus, ``Multi-robot adversarial patrolling: Facing coordinated attacks,'' in International Conference on Autonomous Agents and Multi-agent Systems, 2014, pp. 1093--1100.

\bibitem{rf14} H. H. González-Banos, C.-Y. Lee, and J.-C. Latombe, ``Real-time combinatorial tracking of a target moving unpredictably among obstacles,'' in Robotics and Automation, 2002. Proceedings. ICRA’02. IEEE International Conference on, vol. 2. IEEE, 2002, pp. 1683--1690.

\bibitem{rf15} S. I. Roumeliotis, G. S. Sukhatme, and G. A. Bekey, ``Sensor fault detection and identification in a mobile robot,'' in Intelligent Robots and Systems, 1998. Proceedings., 1998 IEEE/RSJ International Conference on, vol. 3. IEEE, 1998, pp. 1383--1388.

\bibitem{rf16} V. Tzoumas, A. Jadbabaie, and G. J. Pappas, ``Resilient Non-Submodular Maximization over Matroid Constraints,'' arXiv: 1804.01013, 2018.


\bibitem{rf18} B. Schlotfeldt, V. Tzoumas, D. Thakur, and G. J. Pappas, ``Resilient active information gathering with mobile robots,'' IEEE/RSJ International Conference on Intelligent Robots and Systems, 2018.








\bibitem{rap1} R. Yang, C. Kiekintveld, F. Ordóñez, M. Tambe, and R. John, ``Improving resource allocation strategies against human adversaries in security games,'' Artificial Intelligence, vol. 195, pp. 440--469, 2013.

\bibitem{rap2} A. Jiang, Z. Yin, M. Johnson, M. Tambe, C. Kiekintveld, K. Leyton-Brown, T. Sandholm, ``Towards Optimal Patrol Strategies for Fare Inspection in Transit Systems,'' AAAI Spring Symposium, 2012.



\bibitem{rf24} M. Conforti, and G. Cornuéjols, ``Submodular set functions, matroids and the greedy algorithm,'' Discrete Applied Mathematics, vol. 7, no. 3, pp. 251--274, 1984.


\bibitem{rf26} A. Krause, and D. Golovin, ``Submodular function maximization,'' 2014.

\bibitem{grb}  Gurobi Optimization, LLC, ``Gurobi Optimizer Reference Manual,'' 2020,  http://www.gurobi.com


\bibitem{nphard} R. M. Karp, ``Reducibility among combinatorial problems,'' in Complexity of computer computations, 1972, pp. 85--103.


\bibitem{e11} G. Nemhauser, L. Wolsey, and M. Fisher, ``An analysis of approximations for maximizing submodular set functions—I,'' Mathematical programming, vol. 14, pp. 265--294, 1978.







 

\end{thebibliography}
\end{document}